\documentclass{article}
\usepackage{amsmath,amssymb,amsfonts}
\usepackage{graphicx}
\usepackage{hyperref}
\graphicspath{ {./} }
\usepackage{amsmath}
\usepackage{longtable, booktabs}

\title{Machine Translation of Mathematical Text}
\author{Aditya Ohri and Tanya Schmah }
\date{October 9, 2020}

\begin{document}

\maketitle

\begin{abstract}
We have implemented a machine translation system, the PolyMath Translator, 
for {\LaTeX} documents containing mathematical text. The current implementation
translates English {\LaTeX} to French {\LaTeX}, attaining a BLEU score of 53.5
on a held-out test corpus of mathematical sentences.
It produces {\LaTeX} documents that can be compiled to PDF without further editing.
The system first converts the body of an input {\LaTeX} document 
into English sentences containing math tokens, 
using the pandoc universal document converter to
parse {\LaTeX} input. We have trained
a Transformer-based translator model, using OpenNMT, on a combined corpus
containing a small proportion of domain-specific sentences.
Our full system uses both this Transformer model and Google Translate,
the latter being used as a backup
to better handle linguistic features that do not appear in our training dataset.
If the Transformer model does not have confidence in its translation,
as determined by a high perplexity score, then we use Google Translate 
with a custom glossary.
This backup was used 26\% of the time on our test corpus of 
mathematical sentences.
The PolyMath Translator is available as a web service
at \url{www.polymathtrans.ai}.
\end{abstract}

\section{Introduction}

Machine translation for specialized domains such as legal or medical text has received considerable attention. Advances in these areas have been useful in practice and have also given rise to new techniques in areas including domain adaptation \cite{chu2018survey},
automatic term extraction \cite{terryn2019no}
and domain-aware approaches to general-purpose machine translation 
\cite{britz2017effective}. In the present paper we consider the domain of mathematical text, as produced by researchers in mathematics and related fields, and post-secondary teachers of these subjects. To clarify, we consider specialised natural language about mathematics, which we may call informal mathematical writing (even if highly technical), to distinguish it from formal mathematics written in the language of pure logic. In particular, the present paper considers the problem of translation of mathematical writing from English to French. The specific choice of language pair is motivated by the authors’ particular Canadian context. 
The domain of mathematical text has, to our knowledge, not yet been the subject of research in machine translation, beyond some very early work \cite{loh1978machine}\cite{nagao1982english},
although we mention extensive ongoing research in the related areas of mathematical ontology and semantics \cite{wolska2011using} and translation from informal mathematical writing into formal mathematics \cite{wang2020exploration}.
 
Mathematical text presents several features of interest to the researcher. It often mixes natural language with symbolic expressions in the same sentence, with the symbolic expressions having a variety of grammatical roles, including as nouns, pronouns or clauses. Mathematical text has its own grammar and conventions, e.g. “Let x and y be integers” or “Consider the function h = f+g.” (See also \cite{schweiger2008grammar}.) Anecdotally, mathematical writing can be both more and less complex than typical natural language. It is unique in its frequent use of definitions \cite{wolska2011using}, and often complex in its precise description of logical relationships, using vocabulary and grammar that is standard but uncommon outside of mathematics, philosophy and law. While we are not aware of research directly supporting this assertion, it is indirectly supported by Yasseri et al.’s study of Wikipedia entries \cite{yasseri2012practical}, which found that the most complex examples studied, as measured by the Gunning fog index, were those in Philosophy and Physics (Mathematics was not considered). Note that the Gunning fog index depends on mean sentence length and the frequency of long words. On the other hand, there is no suggestion in that paper that the actual vocabulary of mathematical text is larger than usual. On the contrary, anecdotal evidence suggests the opposite. In particular, the authors are aware of many examples of mathematical researchers giving comprehensible technical lectures in languages in which they are not generally fluent. If this simplicity of vocabulary is confirmed, then the domain of mathematical text offers an intriguing possibility of better-than-usual machine translation in this limited domain, while at the same time possessing unique challenges.
 
Practically, there is a great need for machine translation of mathematical text, for researchers, teachers and students. While the commercial possibilities of this domain may not be as obvious as in medicine or law, there is a large base of potential users, and a very large number of documents.
For example the MathSciNet$\textsuperscript{\textregistered}$ database \cite{mathscinet}
contains over 3.6 million items, though not all are in {\LaTeX} format,
and if we expand the definition of mathematical text to include mathematically rich fields such as physics, computer science and engineering, we may consider the 1.8 million papers on the \url{arXiv.org} preprint server, most of which are
in {\LaTeX} format.
In addition, there are millions of students of advanced mathematics with limited fluency in English who rely on English-language textbooks because of the lack of available translations. 
 
One salient obstacle to machine translation of mathematical text has been its symbolic content. The vast majority of mathematical documents are now written in {\LaTeX}, a document-preparation system specific to mathematics, which supports embedded mathematical symbols and expressions, as well as providing document layout features. This ubiquity is at once an obstacle to translation via existing tools (which do not understand {\LaTeX} syntax) and an opportunity, 
since once the {\LaTeX} ``hurdle'' is passed, the vast majority of 
modern mathematical text is now available to machine translation.

The rest of this paper is organised as follows. We outline the custom corpora and glossary
that we use for model training and testing. We then describe our algorithms,
which include {\LaTeX} parsing using pandoc \cite{pandoc} and training a ``Transformer'' neural machine translation  model \cite{vaswani2017attention}.
We evaluate our final system on both whole \LaTeX documents and 
a small test corpus of mathematical sentence pairs.

\section{Methods}\label{sect:methods}

\subsection{Data and Preprocessing}\label{sect:data}

We constructed a small custom glossary of 373 mathematical terms (words and short phrases) by combining several publicly available lists (including \cite{dicoAF,lexiqueAlberta}) and adding a few extra terms.

For all model training, and for most of our validation, we used a combined corpus of pairs of aligned text chunks (mostly sentences), called hereafter our ``main corpus'',
consisting of: a subset of the “Aligned Hansards of the 36th Parliament of Canada” corpus
\cite{hansardcorpus}, a domain-specific subset of the OPUS “Wikipedia” corpus (v1.0) \cite{wolk2014building}, and our own custom ``math'' corpus derived mainly from several research papers of the second author TS. 
The first sub-corpus was
included to provide greater breadth of vocabulary, grammar and style, while the second two focus
specifically on mathematical text.
We describe each of these sub-corpora now.

The “Aligned Hansards of the 36th Parliament of Canada” corpus is a corpus of aligned text chunks (sentences or smaller fragments) extracted from the official records of the 36th Canadian Parliament, including debates from the House of Commons and the Senate\cite{hansardcorpus}. The full corpus consists of 1.28M English-French sentence pairs, containing 33.9M words (English plus French). This is a high-quality corpus consisting mostly of complete sentences. We chose it for its size and quality, and also in the hope that the source material would contain many examples of a formal expository style of language with a structure similar to mathematical text. After removing sentence pairs with irrelevant information such as the title, date, and speaker names, we randomly shuffled the entire filtered corpus and then selected 250,000 sentence pairs.

The OPUS ``Wikipedia'' corpus is a corpus of parallel sentences extracted from Wikipedia by Krzysztof Wołk and Krzysztof Marasek \cite{wolk2014building}, which is part of the OPUS project \cite{tiedemann2012parallel}. The full corpus includes
many language pairs, of which we use only the English/French pair, which consists of 803,670 sentence pairs containing 34M words (English plus French). The subject matter is wide-ranging. In order to focus on mathematical text, we applied a naïve subject matter filter to this corpus: we extracted the only those sentence pairs in which the English sentence contained at least two terms from our custom mathematical glossary described above. We found 16,767 sentence pairs satisfying this criterion. Note that the vocabulary of this reduced dataset contains 55,474 unique tokens, whereas the mean vocabulary size of random subsets of the same corpus, of the same size, was 100,738 (mean over 5 random samples).

The custom math corpus consists of 1,075 sentence pairs in English and French, with the English sentences extracted from several of the second author's  research papers in mathematics.  The English sentences were extracted and translated to French using an early version of our PolyMath translation system with a Google Translate API backend using our custom mathematical glossary. Note that these sentences contains math token words such as ``MATH66X'', which replace mathematical expressions, as described in the next section.

These three sub-corpora were combined into one heterogeneous corpus containing a total of 267,842 English-French parallel text chunks, which we refer to as the “main” corpus in the remainder of this paper. Most of 
the text chunks are sentences, so we refer to them as
``sentences'' hereafter. Each sentence pair in the main corpus was word-tokenized to ensure all tokens in a sentence including punctuation were separated by a space and treated individually during training. 
We randomly shuffled the main corpus, and then randomly split the text pairs into training (80\%), validation (10\%), and test (10\%) sets.

 For additional testing, we used a further custom corpus, the ``linear code'' corpus, containing 160 sentence pairs, extracted from mathematical course notes prepared by M. Nevins on linear error-correcting codes, a subject  that did not appear in our main corpus. Beginning with French sentences extracted from the source document using PolyMath, with mathematical expressions tokenized as explained above, we translated them to English using Google Translate with our custom math glossary, and then proofread them and made manual corrections. 

\subsection{Parsing Methods}

We use a modular design that decouples the {\LaTeX} parsing and machine translation aspects. The first task is to parse the {\LaTeX} document and extract all natural language text for translation while preserving enough document structure and {\LaTeX} commands to reconstruct a full document. The second task, addressed in the next section, is to translate the extracted natural language text to French.

For the first task of parsing the {\LaTeX} document, our main tool was the Pandoc Universal Document Converter \cite{pandoc}, using the Python wrapper \texttt{pypandoc}. For our purposes, since we aim to translate an English {\LaTeX} document to French while preserving the original {\LaTeX} as much as possible, we are ``converting'' a document from {\LaTeX} to {\LaTeX}. Our purpose in doing so is to leverage an intermediate document representation internal to Pandoc, the JSON-formatted 
\textit{abstract syntax tree (AST)}, and a mechanism for performing operations on this syntax tree: pandoc filters, implemented using the Python package \texttt{pandocfilters}.
The abstract syntax tree is organized by block elements, such as paragraphs, bulleted lists, and tables, each of which contains a list of ``inline elements'' including strings of individual words, spaces, and math.
Since the filters integrate into the Pandoc {\LaTeX}-to-{\LaTeX} file conversion, the entire translation process executes with one Pandoc function call. An overview of the process is shown here:

\begin{figure}
\centerline{\includegraphics[scale=0.4]{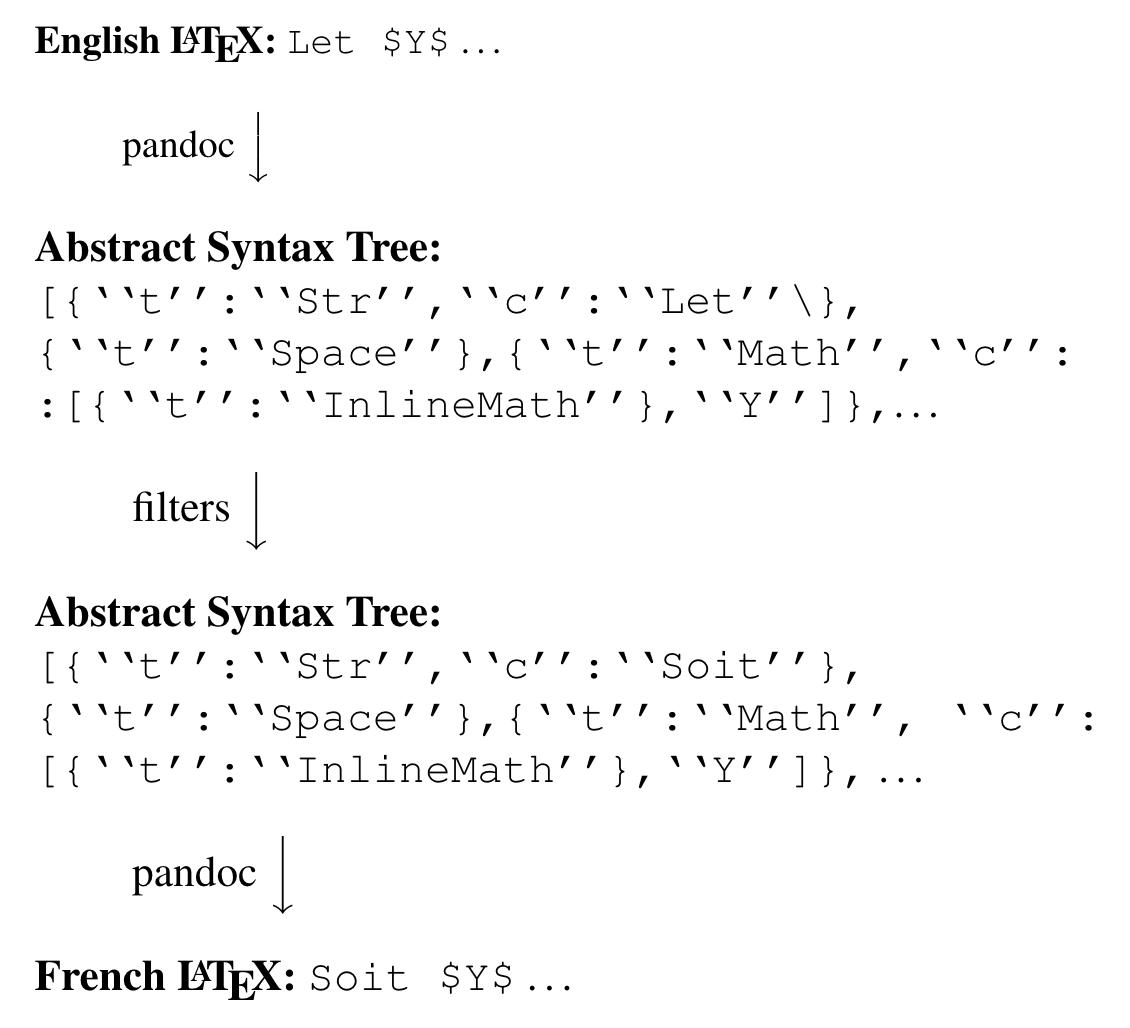}}
\caption{Data flow of \LaTeX{} parsing module using the pandoc universal document converter.
The internal representation is the \textit{abstract syntax tree} (AST) which is in JSON format.}\label{fig:pandoc}
\end{figure}

\begin{figure}
\centerline{\includegraphics[scale=0.5]{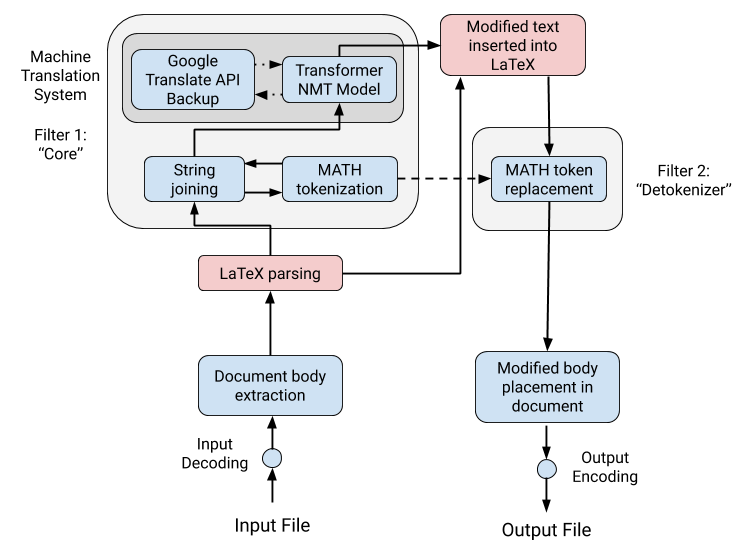}}
\caption{Overview of PolyMath Translator. The pink-coloured modules are implemented using the pandoc universal document converter, see also Fig. \ref{fig:pandoc}.}\label{fig:overview}
\end{figure}

We use two pandoc filters, each of which modifies the abstract syntax tree. 
The first ``core'' Pandoc filter 
translates all block elements that contain natural language text, 
by joining strings of text and math symbols into whole sentences (or sometimes phrases,
as in titles), translating those sentences
(see next section),
and putting the translated sentences back into the abstract syntax tree.
Note that the larger-scaled block-based structure of the document 
is preserved in the abstract syntax tree. 

Typically, pandoc filters act on individual inline elements such as strings. However, for translation of text, this is very limited, as each string containing an individual token would have to be translated separately, instead of a whole sentence. Thus, we combined these inline elements into whole sentences. We accomplished this in a ``string-joining'' function consisting of two layers: manipulating individual block elements through the pandoc interface, and further manipulating the inline elements directly.
Importantly, we include mathematical formulas (inline or displayed) in our sentences,
tokenized into ``MATH'' tokens of the form ``MATHnX'', where n is the index of the token for later retrieval of its corresponding mathematical formula. The original formulas are saved in a JSON object with their corresponding token name as a key. This way, full sentences containing these formulas can be translated without the loss of any important surrounding context. 

For example, consider a simple mathematical sentence in {\LaTeX}: “Let $Y$ have mean $\mu$ and variance $\sigma^2$, and an unknown p.d.f. $p_Y$ that is everywhere nonzero.” Within a JSON-structured paragraph block, this sentence would be represented as a list of inline elements, 
the beginning of which is shown in Fig. \ref{fig:pandoc}.
Without manipulating the mathematical formulas within this sentence, the sentence would be split up by these objects, so the largest possible concatenated phrases for translation would be ``Let'', 
``have mean'', ``and variance'', etc. which would pose a major limitation for translation quality. After tokenizing “math” objects, the whole sentence is concatenated into a single ``Str'' element:
[{``t'':``Str'',``c'':``Let MATH1X have mean MATH2X and variance MATH3X, and an unknown p.d.f. MATH4X that is everywhere nonzero.''}]
The format of the math tokens is such that they are treated as unknown English words by
the translation module(s), and left unchanged in the translated sentences.
Mathematical expressions can now be treated as individual tokens during translation, and the entire sentence can be translated, optimizing translation quality. Note that after the filter runs, pandoc automatically converts the ``Str'' sentence object back into individual ``Str'' word and ``Space'' objects, as in the original abstract syntax tree. 

Once all natural language text in the modified {\LaTeX} document has been translated to French,
it remains to replace the ``MATH'' tokens with the original mathematical formulas,
using the saved JSON object created by the ``core'' pandoc filter.
This is accomplished by
a second pandoc filter called the ``detokenizer''.

Finally, pandoc creates a new \LaTeX{} file using the translated text and the structure of the original document.
The entire process is represented in Fig. \ref{fig:overview}.
Note that some changes to the \LaTeX{} commands are introduced as a result of pandoc's 
abstract syntax tree not being able to completely represent all \LaTeX{} commands.
However when the original and translate documents are compiled to PDF format, 
very few differences in format are seen.

\subsection{Translation methods}
\label{sect:translation}

For translating text, we trained a custom neural machine translation (NMT) model using the Transformer architecture for neural sequence transduction introduced by Vaswani et al. \cite{vaswani2017attention}.
The Transformer is a neural sequence transduction model, i.e. a 
``sequence-to-sequence'' translator implemented using a neural network
that outputs, for any position $t$ in the output sequence $y$,
a conditional distribution $p\left(y_t \big| y_{<t}, x\right)$
based on the entire input sequence $x$ and the preceding outputs
$y_{<t}$.
Like most other such models, the Transformer has an encoder-decoder
structure. 
\textit{Attention} mechanisms apply weights to elements of
the input sequence, which vary according to the position $t$ in the
output sequence. This allows the network to ``pay attention'' to
certain inputs, for example when producing the first word of
an output sentence, the network may pay most attention to the first
word of the input sentence. 
The key distinguishing feature of the Transformer model is its use of Multi-Head Self-Attention which allows it view the input sequence
from different ``points of view'' by applying several parallel attention functions. For example, when encoding the word ``kicked'' in the sentence ``I kicked the ball'', one may pay more attention to the ``I'' or the ``ball'', corresponding to the asking the questions “who performed the action?” or  “what was kicked?”. In this toy example, the first point of view would
aid most in conjugating the corresponding output verb, while the second
point of view would aid most in translating the verb root.
A key feature of the Transformer attention mechanism is that it
is highly parallelizable, allowing much faster training than previous
comparable models.
We trained a Transformer ``base model'' as implemented in
OpenNMT \cite{klein-etal-2017-opennmt}, using the training subset of the main corpus described above. Training details are given in the next section.

The output of our trained Transformer model is already high-quality,
as will be seen in the next section. However, while we can expect our 
model to have good performance on mathematical text, thanks to 
the deliberate inclusion of mathematical text in our training set,
we cannot expect it to perform as well on general English text
as commercial translation services such as Google Translate,
due to our limited training set and off-the-shelf Transformer architecture.
Even mathematical sentences may contain linguistic features not
seen in our limited training set. For this reason, we used Google Translate as a ``backup'' translator
in our final system, as follows.

We first run all sentences through our main Transformer model.
The output of this process is not just a translated sentence
but also a cumulative log conditional likelihood score
$\sum_t \log p\left(y_t \big| y_{<t}, x\right)$, where the sum is over all tokens
in the output sentence. 
Dividing this by the length of the sentence gives the mean 
log conditional likelihood per token, which is a measure of how 
confident the system is of its prediction.
This is commonly converted into a \textit{perplexity} value,
calculated as $\exp\left(-\textrm{mean conditional likelihood}\right)$, which is 
lowest for the most confident predictions.
By visual inspection of the quality of translations of validation sentences, we established a threshold value of $2.05$ for perplexity,
below which we were satisfied with the quality of translation.
At test time, whenever our Transformer model produced a sentence
with perplexity above this threshold, we discarded the output
and instead re-translated the input sentence using the 
Google Translate API with our custom math glossary.

\subsection{French-specific \LaTeX{} modifications}

Finally, since French {\LaTeX} has its own typographical conventions, PolyMath adds a ``french'' option to the document class and adds the following lines to the document header:
\begin{verbatim}
\usepackage[T1]{fontenc}
\usepackage{babel}
\end{verbatim}
It also explicitly translates all {\LaTeX}-style double quotes (i.e.
two consecutive backquotes or two consecutive single quotes)
into \verb|\og| and \verb|fg{}| respectively.

\subsection{Training, Validation and Testing}

We used the BLEU metric \cite{papineni2002bleu} to evaluate the quality of machine translations as compared to given target translation.
This metric is standard in the field and has been shown to correlate well with human judgement of translation quality.

As noted earlier, our main corpus was randomly split into training (80\%), validation (10\%), and test (10\%) sets.
The training subset contains 214,272 English-French sentence pairs,
while the validation and test subsets each contain 26,785 sentence pairs.

We trained the Transformer model with a vocabulary of 50,000 words on the training subset of the main corpus. We set the batch size to 3072 tokens and a maximum of 100,000 steps or 65 epochs of the entire corpus for training, and we used default values of all other hyperparameters.
We did not tune any model hyperparameters.
We used an early stopping condition, to avoid over-fitting and save computational cost once the model converges. for which we used the validation subset of the main corpus. The stopping condition was:
stop when the validation BLEU score does not improve by more than 0.2 points over the last 4 evaluations. After 50,000 steps or 32 epochs, our model met the early stopping criteria and stopped training with a validation BLEU score of 28.7.
This training required 12 hours on 8 Tesla V100 GPUs.

During runtime, the Transformer model first infers the translation of a particular sentence, and if it has a test perplexity of less than or equal to 2.05, this prediction is used. Otherwise, the sentence is translated by Google Translate with our custom math glossary,
as discussed in Section \ref{sect:translation}.

After training the NMT model and integrating the other components described above, we used the testing corpora to evaluate the quality of the PolyMath system.
We compare the entire PolyMath Translation system with its two components: the NMT model alone; and Google Translate alone.
For each system, a test BLEU score for was computed for two corpora: 
the testing subset of the main multi-domain corpus; and the ``linear block'' corpus
of mathematical text.

\section{Results}

Table \ref{tab:bleu} shows our main experimental results, which are
BLEU scores calculated on our two test corpora,
for translations made by: the PolyMath Translator (full system);
Google Translate only (with no glossary); 
and our trained Transformer NMT model only.
As expected, the PolyMath full system produces better results than
the two translation systems that it uses.
It is also noteworthy that the Transformer NMT model
outperformed Google Translate, even on multi-domain text.
This may be partly due to the 
similarity of training and test data, which gave the Transformer model
an advantage. However it does suggest that our main corpus
was large enough to support a comprehensive language model.

The highlight of these BLEU results is the score of 53.5
on mathematical text (the ``linear code'' corpus).
This is much higher than the state-of-the-art for 
general English to French machine translation, which is 41.8,
as attained by the ``big'' Transformer model \cite{vaswani2017attention}.
We note that during translation of the ``linear code'' corpus,
the PolyMath system used the Transformer NMT model for 74\% of the sentences,
and used the ``backup system'' of Google Translate with custom glossary for the other 26\% of the sentences.

\vspace{0.5cm}

\begin{table}[ht]
    \centering
    \begin{tabular}{|c|ccc|}
    \hline
         & Google & Transformer& PolyMath \\
    \hline
    multi-domain text     & 27.7 & 29.0 & \textbf{32.1} \\
    mathematics text & 46.5 & 50.4 & \textbf{53.5} \\
    \hline
    \end{tabular}
    \caption{Test results on two held-out corpora: BLEU scores for (i) the ``test'' subset of our multi-domain main corpus (top row); and (ii) the ``linear code'' corpus consisting entirely of mathematical text (bottom row).
    The columns correspond to: (i) Google Translate;
    (ii) the Transformer model trained in this paper;
    (iii) then entire PolyMath Translator system.}
    \label{tab:bleu}
\end{table}

\begin{table*}[ht]
    \centering
    \small
    \begin{tabular}{|c|l|}
\hline
Original English {\LaTeX}
: & \includegraphics[scale=0.22]{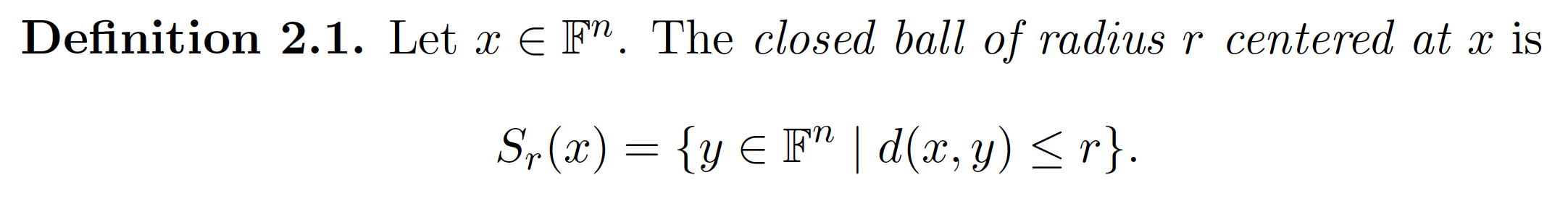} \\
    \hline
& \\
    Google Translate: & \begin{minipage}
    {0.5\textwidth}
\verb|\ begin {defn}|\\
\verb|Soit $ x \ dans \ F ^ n$.}|\\
\verb|La \ define {boule fermée de rayon $ r $ |\\
\verb|centrée sur $ x $} est|\\
\verb|$$|\\
\verb|S_r (x) = \ {y \ in \ F ^ n \ mid d (x, y) \ leq r \}.|\\
\verb|$$|\\
\verb|\ end {defn}|  \\  
\end{minipage} \\
%& \\
\hline
    PolyMath output {\LaTeX}: & \includegraphics[scale=0.5]{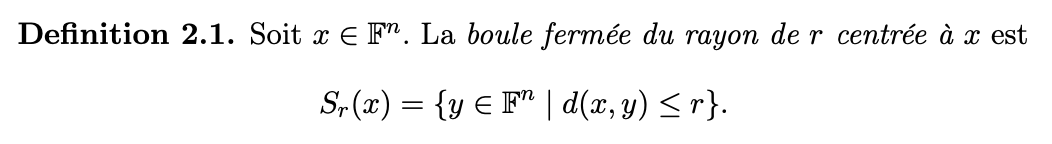} \\
    \hline
    \end{tabular}
        \caption{Illustration of two translations of English {\LaTeX} into French.
    The top row shows an excerpt from a compiled {\LaTeX} document. The 
    corresponding {\LaTeX} source code
    was entered into to two English-to-French translation systems:
    (i) Google Translate correctly translates the words, however is unaware of {\LaTeX} syntax,
    resulting in an un-compilable document (e.g. ``$\backslash$in'' is translated to ``$\backslash$ dans''); 
    while (ii) PolyMath correctly translates the entire {\LaTeX} document into
    a French {\LaTeX} document that compiles without further editing.}

    \label{tab:texexcerpt}
\end{table*}

Finally, in Table \ref{tab:texexcerpt} we illustrate that the Polymath
Translator is a complete {\LaTeX} document translation system, integrating natural language
translation with LaTeX document parsing and French language support.

All of the {\LaTeX} source documents used in this study, as well as the present manuscript, 
were translated to French by the PolyMath Translator system, and the results compiled to PDF using TeXShop or Overleaf. 
Most of the output documents compiled to PDF without any manual editing required. 
The translated version of the present manuscript is included as an ancillary file, compiled to PDF; 
this is the output of the PolyMath Translator, with minor manual edits made to only 6 lines of the output file
(mostly related to figures).
There are some slight differences in format between the original
manuscript and the French version, due to the pandoc internal 
representation (the abstract syntax tree) not being a perfect representation of 
the {\LaTeX} input. However the 
formatting is very similar to the original paper, and the 
English to French translation is very good.

\section{Discussion}
We have implemented a prototype machine translation system for {\LaTeX}
documents containing mathematical text. The current implementation
translates English {\LaTeX} to French {\LaTeX}, attaining a BLEU score of 53.5
on a held-out test corpus of mathematical sentences.
Our main contributions are: (i) an algorithm for parsing {\LaTeX}
documents into sentences containing math tokens (using pandoc)
and interfacing with existing natural language translators;
(ii) a Transformer-based translator model trained (using OpenNMT) on a heterogeneous corpus
containing a small proportion of domain-specific sentences;
(iii) the PolyMath Translator system, which incorporates the previous two elements and
uses Google Translate with a custom glossary as a backup.
This system is available as a web-service at
\url{polymathtrans.ai}.

In most of our experiments, we found that 
the output of PolyMath compiled to PDF without error or manual editing; thus PolyMath
is robust in its handling of {\LaTeX} syntax.
However PolyMath depends on the pandoc converter, which cannot represent all aspects of \LaTeX{} and its customizations (e.g. new commands),
and so the output of PolyMath always changes the \LaTeX source somewhat, more so when many customizations are used.
Thus we expect PolyMath to be most useful with documents in simpler, standard formats.

Our BLEU score of 53.5 on mathematical text is much higher than  
the state-of-the-art for 
general English to French machine translation, which is 41.8,
attained by the ``big'' Transformer model \cite{vaswani2017attention}.
However we achieved a lower BLEU score of 32.1 on our 
main multi-domain corpus (29.0 using the Transformer model only), which is more in line with expectation,
especially since we 
used the ``base model'' Transformer and only trained for 12 hours on 8 V100s,
while the state-of-the-art result of 41.8 required
training a ``big'' Transformer model for 3 days on 8 V100s.

Comparing our BLEU results for the two testing corpora: 
32.1 for multi-domain text vs.
53.5 on mathematical text;
we conclude that, since we used the same translation system,
our increased score on mathematical text is
due to the relative simplicity of this domain.
Indeed, while ``simplicity'' has many aspects, we have verified
one important aspect: the vocabulary of our subset of mathematical sentences from the OPUS Wikipedia corpus was approximately half 
of the vocabulary of other subsets of the same corpus of the same size,
as noted in Section \ref{sect:data}.
This supports the anecdotal evidence, and the observations in \cite{yasseri2012practical}, that were noted in the Introduction.
It would be interesting to study this further, and compare the mathematical domain with other specialised
domains in this respect.
Since mathematical text has a smaller vocabulary than general natural language, it offers an intriguing possibility of better-than-usual machine translation, and other natural language processing, in this limited domain, as suggested by our high
BLEU score of 53.5 on a test corpus, using a prototype translation system that has obvious room for improvement.

We intend to further develop this system. Several obvious opportunities
exist for improvement. 
We can expand and improve our corpora and glossary, using both automated tools, for example automatic multi-word terminology extraction \cite{wolk2014building,sandrih_krstev_stankovic_2020}, and manual proofreading.
This might involve the use of large document collections including 
\url{arXiv}, which includes some publicly available \LaTeX{} sources, see \cite{clement2019use}.
We can incorporate recent advances in machine translation research
in areas including deep learning \cite{zhang2020neural, popel2020transforming},
Bayesian modelling \cite{yu2020better}, and
methods for incorporating whole document context \cite{mace2019using}.

Also, we will investigate alternative methods of \LaTeX{} parsing that will allow us
to exactly retain all of the \LaTeX{} commands in the original document, rather than 
passing them through the pandoc internal representation, which introduces some changes as mentioned above.

An intriguing possibility specific to the domain of mathematical text
is to improve translations by semantic understanding of the
content of mathematical formulas, see e.g.
\cite{kristianto2014extracting}. A simpler early target would be
to classify the formulas by their parts of speech, usually nouns, pronouns or clauses; and encode them in a way that is usable by the translation
modules.

We intend to support other language pairs in the near future.
Further, given the relative ease of machine translation
in the mathematical text domain, as noted above,
and at the same time, the lack of curated corpora of 
mathematical sentence pairs, we are optimistic that semi-supervised
translation \cite{cheng2019semi} and multi-language models \cite{firat2016multi}
will be successful in this domain.

\end{document}